\documentclass[10pt,twocolumn,letterpaper]{article}

\usepackage{iccv}              

\definecolor{iccvblue}{rgb}{0.21,0.49,0.74}
\usepackage[pagebackref,breaklinks,colorlinks,allcolors=iccvblue]{hyperref}

\def\paperID{5510} 
\def\confName{ICCV}
\def\confYear{2025}

\usepackage{epstopdf}
\usepackage{graphicx}
\usepackage{amsmath}
\usepackage{amsbsy}
\usepackage{amssymb}
\usepackage{booktabs}

\usepackage{multirow}
\usepackage{arydshln}
\usepackage{indentfirst}

\usepackage{algorithm, algorithmic}
\usepackage{pifont}
\usepackage{bbding}
\usepackage{gensymb}
\usepackage{colortbl}
\usepackage{bbm}

\title{Fusion Meets Diverse Conditions: A High-diversity Benchmark and Baseline for UAV-based Multimodal Object Detection with Condition Cues}
\author{Chen Chen \quad Kangcheng Bin\footnotemark[1] \quad Ting Hu \quad Jiahao Qi \quad Xingyue Liu\\  \quad Tianpeng Liu \quad Zhen Liu \quad Yongxiang Liu\footnotemark[1] \quad Ping Zhong\footnotemark[1]\\
National University of Defense Technology, China\\
{\tt\small $\{$chenchen21c, binkc21, huting, qijiahao1996, liuxingyue18$\}$@nudt.edu.cn,}\\
{\tt\small $\{$liutianpeng2004, zhen\_liu, zhongping$\}$@nudt.edu.cn, lyx\_bible@sina.com}
}

\begin{document}
\twocolumn[{%
	\renewcommand\twocolumn[1][]{#1}%
	\maketitle
	\begin{center}
        \vspace{-1.0cm} 
		\centering
		\captionsetup{type=figure}
		\includegraphics[width=0.99\textwidth]{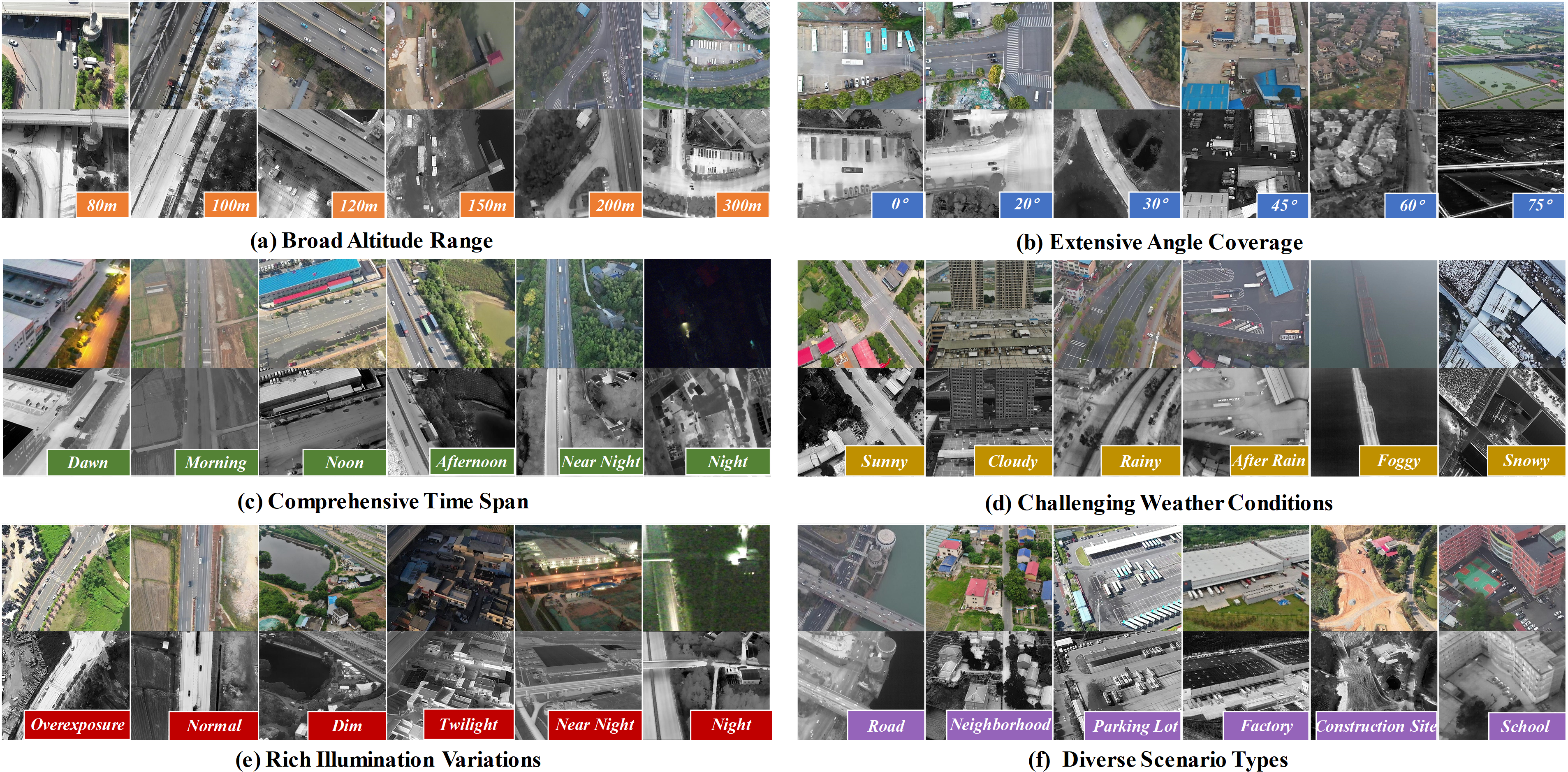}
        \vspace{-0.3cm}
		\captionof{figure}{High-diversity imaging conditions in our ATR-UMOD. Some representative examples are shown for each condition.
		  \textbf{(a) Broad Altitude Range:} It encompasses an altitude range from 80m to 300m, offering a rich resource for multi-scale object analysis.
		  \textbf{(b) Extensive Angle Coverage:} Nearly full angular coverage from 0° to 75° ensures comprehensive object appearances from various viewpoints.
		  \textbf{(c) Comprehensive Time Span:} All-day data collection captures fluctuations in light, shadows, and thermal characteristics over time.
		  \textbf{(d) Challenging Weather Conditions:} Incorporating 7 typical and extreme weather conditions enhances robustness in real-world applications.
		  \textbf{(e) Rich Illumination Variations:} It covers 6 illumination levels from lightless to high-light, improving adaptability to varying image qualities.
		  \textbf{(f) Diverse Scenario Types:} Considering cross-scene generalization, it spans 11 scenarios types with complex backgrounds.
		  These conditions are additionally annotated in each image pair, providing valuable high-level contextual insights and establishing a comprehensive benchmark for condition-specific performance evaluation.
		}
		\label{fig:1}
		\vspace{-0.3cm}
	\end{center}%
}]

\maketitle

\renewcommand{\thefootnote}{\fnsymbol{footnote}} 
\footnotetext[1]{Corresponding authors.} 

\begin{abstract}
    Unmanned aerial vehicles (UAV)-based object detection with visible (RGB) and infrared (IR) images facilitates robust around-the-clock detection, driven by advancements in deep learning techniques and the availability of high-quality dataset.
    However, the existing dataset struggles to fully capture real-world complexity for limited imaging conditions.
    To this end, we introduce a high-diversity dataset ATR-UMOD covering varying scenarios, spanning altitudes from 80m to 300m, angles from 0° to 75°, and all-day, all-year time variations in rich weather and illumination conditions.
    Moreover, each RGB-IR image pair is annotated with 6 condition attributes, offering valuable high-level contextual information.
    To meet the challenge raised by such diverse conditions, we propose a novel prompt-guided condition-aware dynamic fusion (PCDF) to adaptively reassign multimodal contributions by leveraging annotated condition cues. 
    By encoding imaging conditions as text prompts, PCDF effectively models the relationship between conditions and multimodal contributions through a task-specific soft-gating transformation.
    A prompt-guided condition-decoupling module further ensures the availability in practice without condition annotations.
    Experiments on ATR-UMOD dataset reveal the effectiveness of PCDF.
\end{abstract}   
\vspace{-0.5cm} 
\section{Introduction}
\label{sec:intro}


Unmanned aerial vehicle (UAV)-based object detection using visible (RGB) and infrared (IR) images (referred to as RGB-IR UOD) offers a promising solution for trafﬁc monitoring, military reconnaissance, and so on~\cite{DBLP:conf/icra/BozcanK20,DBLP:conf/aipr/ChenHWOARFCRW22,DBLP:journals/aei/JiSWXYM25,DBLP:journals/aeog/MaLJZWJ23}.
Its advancement heavily depends on comprehensive datasets, as modern computer vision techniques predominantly rely on a data-driven manner.
DroneVehicle~\cite{DBLP:journals/tcsv/SunCZH22}, the pioneer dataset for RGB-IR UOD, holds significant potential to facilitate progress in this field.
However, it is constrained by a narrow variety of imaging conditions in altitude, angle, time, weather, illumination, and scenario, which struggles to fully represent the complexity in real-world scenarios.

\begin{figure}[t]
  \centering
   \includegraphics[width=1.0\linewidth]{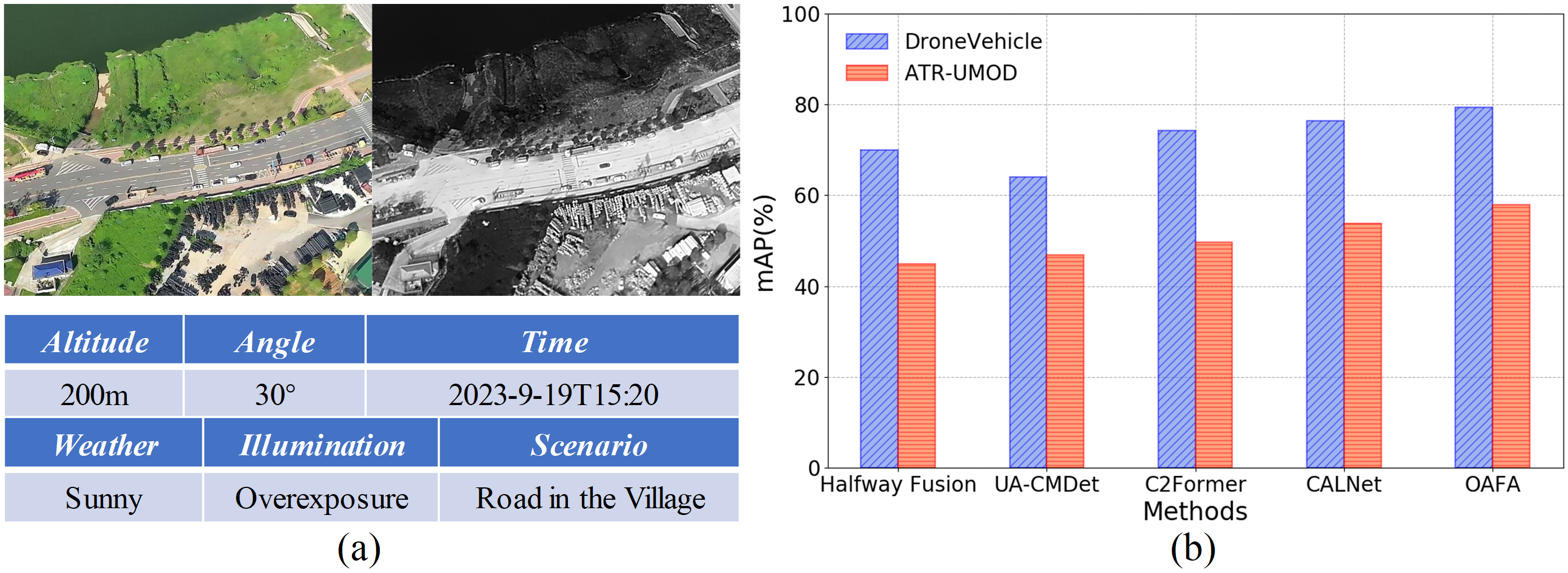}
   \vspace{-0.85cm}
   \caption{Advantage and challenge in our dataset. (a) Annotation example. (b) Performance degradation on ATR-UMOD.}
   \label{fig:2}
   \vspace{-0.8cm}
\end{figure}

To address this issue, we introduce ATR-UMOD, a novel dataset to provide more comprehensive data support for RGB-IR UOD and improve model robustness against complex real-world conditions.
Compared to the existing dataset, it excels in some aspects:
(1) \textbf{Diverse imaging conditions.} As illustrated in \cref{fig:1}, it was built at flight altitudes ranging from 80m to 300m and camera angles from 0° to 75° covering all-day and all-year conditions.  
It also spans a wide variety of scenarios with richer weather and illumination variations, closely mirroring real-world complexities.
(2) \textbf{Richer object types.}
We provide 11 fine-grained object categories covering typical objects in real-world applications, supporting fine-grained detection from UAV perspectives.
(3) \textbf{Extra condition annotations.}
We additionally annotated 6 condition attributes for each image pair, as indicated in \cref{fig:2}\textcolor{iccvblue}{a}, providing valuable high-level contextual insights and
establishing a comprehensive benchmark for condition-sensitive performance evaluation.


ATR-UMOD captures the complexity of real-world conditions, but it also introduces new challenges. As shown in \cref{fig:2}\textcolor{iccvblue}{b}, most existing methods underperform on ATR-UMOD, likely due to visual information bottlenecks in such complex conditions~\cite{DBLP:conf/icml/ZhaoDBCZZQC0WG24}. 
To this end, several studies have explored imaging condition cues, such as illumination, as auxiliary information~\cite{,DBLP:conf/eccv/ZhouCC20,DBLP:journals/tim/ZhangYHWY23}. 
Motivated by this, we try to \textbf{leverage conditions as auxiliary contextual prompts} for improved detection performance across diverse conditions.

Studies in this area dynamically reassigned multimodal contributions based on imaging conditions for trustworthy fusion~\cite{DBLP:conf/icml/ZhangWZHFZP23,DBLP:conf/aaai/ZhengTWH023}.
They modeled the relationships between condition representations and multimodal contributions for dynamic fusion~\cite{DBLP:journals/inffus/GuanCYCY19,DBLP:journals/pr/LiSTT19,DBLP:journals/remotesensing/WuSWTZS23}, enhancing effective information utilization from high-contribution modaliies while mitigating noises from subordinate ones.
Despite these advances, two challenges still remain:
(1) \textbf{Inadequate Condition Representation.}
They often focused solely on single condition attribute (\eg, illumination), neglecting others which also impact multimodal reliability~\cite{DBLP:journals/corr/abs-2410-10791}.
Furthermore, condition representations are typically derived from a condition prediction model~\cite{DBLP:journals/pr/LiSTT19,DBLP:journals/remotesensing/WuSWTZS23}.
In this situation, diverse condition representations require advanced multi-label prediction techniques, which are challenging due to the diversity and interdependence of condition attributes~\cite{DBLP:journals/pr/ReadMH17}.
(2) \textbf{Task-irrelevant Condition-guided Pipeline.}
Existing methods often depend on pretext tasks to model the relationships between conditions and multimodal contributions, such as utilizing an illumination prediction task to assign illumination values as RGB contributions~\cite{DBLP:journals/inffus/GuanCYCY19}.
The mismatch of the optimization objectives between pretext tasks and the detection task result in suboptimal multimodal contributions and ultimately compromise performance.

\begin{table*}[hbt!]
  \centering 
  \scriptsize
  \setlength{\tabcolsep}{3.2mm}{
  \resizebox{0.99\linewidth}{!}{
  \begin{tabular}{|c|c|c|c|c|c|c|c|c|c|c}
  \hline
  \multirow{2}{*}{\makebox[0.02\textwidth][c]{Dataset}} 
   & \multirow{2}{*}{\makebox[0.028\textwidth][c]{Categories} }   & \multicolumn{6}{c|}{\makebox[0.01\textwidth][c]{Conditions}} & \multirow{2}{*}{\makebox[0.07\textwidth][c]{\begin{tabular}[c]{@{}c@{}}Conditions \\ Labeled \end{tabular}}}
  & \multirow{2}{*}{\makebox[0.001\textwidth][c]{Pubilsh}}  \\ \cline{3-8} 
   &   & \makebox[0.01\textwidth][c]{Altitude} & Angle & Time & Weather &Illumination  & \makebox[0.01\textwidth][c]{Scenario} &  & \\ 
  \hline 
  \hline

  DroneVehicle   & 5  & \begin{tabular}[c]{@{}c@{}}80m\\ 100m\\120m\end{tabular}&\begin{tabular}[c]{@{}c@{}}15\degree\\30\degree\\45\degree\end{tabular}   & \begin{tabular}[c]{@{}c@{}}Morning \\Afternoon \\Night\end{tabular}  & \begin{tabular}[c]{@{}c@{}}Sunny\\Cloudy\\  Foggy\\Night\end{tabular} & \begin{tabular}[c]{@{}c@{}}Day\\Night\\Darknight\end{tabular}& Urban & \ding{56} & TCSVT 2022 \\

  \hline
  ATR-UMOD   & 11   & \begin{tabular}[c]{@{}c@{}}80m $\backsim$ 300m\end{tabular} &\begin{tabular}[c]{@{}c@{}}0\degree $\backsim$ 75\degree\end{tabular}   & \begin{tabular}[c]{@{}c@{}}Dawn \\ Morning \\Noon \\Afternoon\\Near Night\\Night\end{tabular}& \begin{tabular}[c]{@{}c@{}}Sunny\\ Cloudy\\Rainy\\After Rain\\ Snowy\\ Foggy\\Night\end{tabular} & \begin{tabular}[c]{@{}c@{}}Overexposure\\ Normal\\  Dim\\Twilight\\ Near Night\\ Night\end{tabular} & \begin{tabular}[c]{@{}c@{}}Urban\\Suburban\\Village\end{tabular} & \ding{52} & ICCV 2025 \\

  \hline
  
  \end{tabular}}}
  \vspace{-0.3cm}
  \caption{Comparison with the existing RGB-IR UOD dataset. 
  \label{tab:1}}
  \vspace{-0.8cm}
\end{table*}




To this end, we propose \textbf{P}rompt-guided \textbf{C}ondition-aware \textbf{D}ynamic \textbf{F}usion (\textbf{PCDF}), a novel method that adaptively reassigns multimodal contributions based on condition prompts, improving detection robustness across diverse conditions.
Leveraging CLIP\textquotesingle s powerful text-semantic representation capability~\cite{DBLP:conf/icml/RadfordKHRGASAM21}, we encode multi-label conditions as text prompts to obtain expressive and robust condition representations.
Considering the varying sensitivity of multimodal contributions to different condition attributes in each sample, a sample-specific condition prompt learning (SCPL) strategy is adopted to ensure relevant prompt construction.
To establish task-specific relationships between conditions and multimodal contributions, we introduce a condition-aware dynamic fusion (CDF) module that refines feature reweighting through a detection-oriented normalized soft-gating transformation.
Additionally, since explicit condition labels are unseen in practices, we design a prompt-guided condition-decoupling (PCD) module where condition-specific features generate prompts to dynamically modulate condition-invariant features.
Extensive experiments on the ATR-UMOD dataset validate the effectiveness and robustness of PCDF under diverse conditions.
\vspace{-0.3cm}
\section{Related Work}
\label{sec:formatting}

\subsection{RGB-IR UOD Dataset}
RGB-IR UOD is a promising and emerging field, yet its datasets remain scarce with DroneVehicle~\cite{DBLP:journals/tcsv/SunCZH22} which has been instrumental in advancing research.
Despite significant contributions, its imaging conditions are restricted by fixed flying altitudes and camera angles, limited imaging time, exclusive clear weathers, restricted illumination variations, and simple scenarios, which cannot fully capture the dynamic changes in object scales, viewpoints, and appearances, as well as the complexity of backgrounds.
Additionally, only 5 object categories limits the range of potential applications and undermines the generalization ability of detection model.
Finally, the lack of condition annotations prevents the exploration of conditional impacts on multimodal fusion and hinders comprehensive evaluation under diverse conditions.
To address these issues, our dataset features annotated 11 object categories and 6 additional condition attributes, covering a broader range of imaging conditions across multiple dimensions, as detailed in \cref{tab:1}, which better mirrors real-world complexities and provides a comprehensive benchmark for condition-sensitive, fine-grained detection from UAV perspectives.

\subsection{Condition Representation Method}
Leveraging condition representation as additional information has proven effective in computer vision tasks~\cite{DBLP:conf/iccvw/ChuPWHSBLA19,DBLP:conf/cvpr/YangLSZTZLY22,DBLP:journals/tgrs/LuoHZZKJ24,DBLP:conf/iccv/AodhaCP19,DBLP:journals/inffus/GuanCYCY19,DBLP:journals/remotesensing/WuSWTZS23}.
For example, Chu \etal ~\cite{DBLP:conf/iccvw/ChuPWHSBLA19} pioneered use a fully connected network to model geolocation representations for fine-grained classification, yet it may fail to capture rich condition semantics due to the lack of explicit constraints.
To solve this, Guan \etal~\cite{DBLP:journals/inffus/GuanCYCY19} extracted illumination representations from a Day-Night prediction network for RGB-IR fusion.
Wu \etal~\cite{DBLP:journals/remotesensing/WuSWTZS23} introduced region-wise illumination prediction for finer representations.
However, they only focus on a single condition, ignoring other effective condition attributes.
Moreover, multi-condition representations with such prediction networks remain challenging due to the diversity and interdependence of the condition attributes.
To this end, we propose a multi-condition-guided fusion method, leveraging CLIP\textquotesingle s robust and flexible semantic representations ability to encode multi-conditions as text prompts for effective condition representations.

\subsection{Condition-guided Fusion Method}
Since imaging conditions greatly affect multimodal reliabilities (\eg, IR outperforms RGB in low-light conditions)~\cite{DBLP:journals/corr/abs-2404-18947}, condition-guided fusion methods have gained increasing attention~\cite{DBLP:journals/inffus/GuanCYCY19,DBLP:journals/tim/ZhangYHWY23,DBLP:journals/pr/LiSTT19,DBLP:journals/kbs/ChenLZ23}.
They aimed to dynamically reassign multimodal contributions based on condition-sensitive modality reliability for trustworthy fusion.
Guan \etal~\cite{DBLP:journals/inffus/GuanCYCY19} pioneered illumination-guided fusion by a Day-Night prediction network and directly treated Day probabilities as RGB reliabilities.
To prevent modality imbalance under extreme illuminations, Zhang \etal ~\cite{DBLP:journals/tim/ZhangYHWY23} introduced a linear gate function to optimize reliability.
IAF R-CNN~\cite{DBLP:journals/pr/LiSTT19} and IGT~\cite{DBLP:journals/kbs/ChenLZ23} further modeled nonlinear reliabilities with a Sigmoid function.
However, all of them rely on condition prediction tasks that are misaligned with detection objectives, leading to suboptimal modality reliabilities.
In contrast, we propose a detection-oriented soft-gating transformation that leverages rich-semantic condition representations to learn task-specific multimodal reliability.

\vspace{-0.2cm}
\section{ATR-UMOD Dataset}
\begin{figure*}[htb!]
  \centering
   \includegraphics[width=0.99\linewidth]{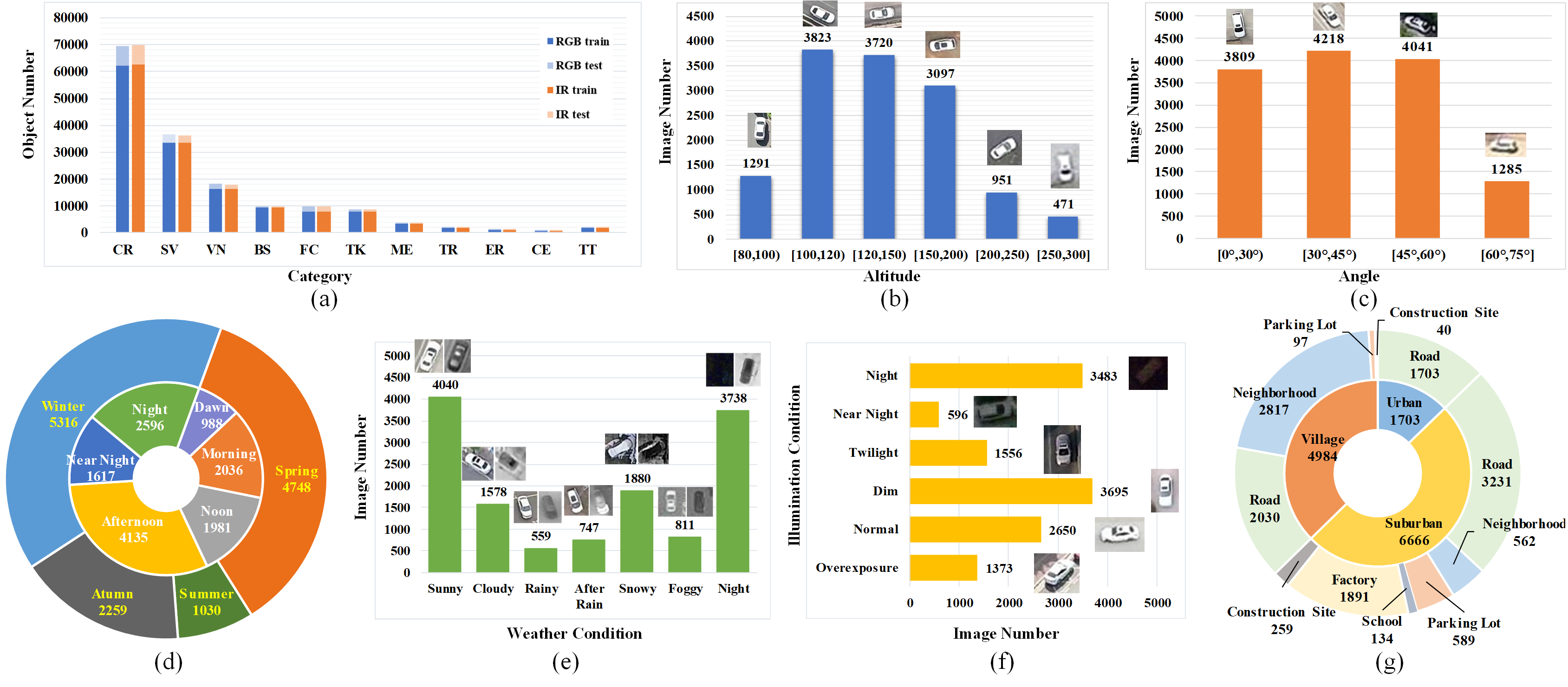}
   \vspace{-0.5cm}
   \caption{The object and attribute statistics of the ATR-UMOD dataset. Note that CR, SV, VN, BS, FC, TK, ME, TR, ER, CE  and TT represent car, SUV, van, bus, freight car, truck, motorcycle, trailer, excavator, crane, and tank truck categories, respectively.}
   \label{fig:3} 
   \vspace{-0.7cm}
\end{figure*}
\label{sec:3}

\vspace{-0.1cm}
\subsection{Dataset Construction}
\textbf{Data collection and object annotation.}
ATR-UMOD is built spanning diverse imaging conditions in flying altitude, camera angle, shooting time, weather, illumination and scenario.
Due to hardware limitations, raw RGB-IR images suffer from inevitable cross-modal misalignment for differences in imaging space and time~\cite{DBLP:conf/eccv/YuanWW22}.
To this end, we employed homography transformation~\cite{DBLP:journals/pami/Zhang00} and region cropping for spatial calibration and timestamp alignment for temporal calibration.
For annotation, RGB and IR objects were labeled separately with oriented bounding boxes.

\textbf{Attribute annotation.}
We enriched ATR-UMOD with detailed condition annotations, offering essential context to address visual bottlenecks and facilitate in-depth analysis of conditional impact on multimodal fusion.
Specifically, we labeled 6 key condition attributes for each image pair: \textit{Altitude}, \textit{Angle}, \textit{Time}, \textit{Weather}, \textit{Illumination}, and \textit{Scenario}.

\textbf{Training and testing sets.}
It is divided into training and testing sets with 11,850 and 1,503 image pairs, respectively.
To ensure rigorous evaluation, the subsets are derived from non-overlapping scenarios.
Additionally, as shown in \cref{fig:3}\textcolor{iccvblue}{a}, the object distribution across each subset has been carefully balanced to minimize data bias.

\vspace{-0.1cm}
\subsection{Dataset Statistics}
\textbf{Object statistics.}
It contains 13,353 well-aligned RGB-IR image pairs at $640\times 512$ resolution, covering 161,799 RGB objects and 162,253 IR objects across 11 categories.
As depicted in \cref{fig:3}\textcolor{iccvblue}{a}, it exhibits a pronounced long-tail distribution~\cite{DBLP:journals/pami/ZhangKHYF23}, with the car being the dominant category. 
This distribution closely reflects real-world situations but also introduces significant challenges for detection models.

\textbf{Altitude statistics.}
Flying altitude of UAV significantly affects the object scales.
According to \cref{fig:3}\textcolor{iccvblue}{b}, the dataset spans altitudes from 80m to 300m, capturing substantial scale variations.
This broad range of altitudes promotes detection generalization across different object scales.

\textbf{Angle statistics.} Angle is the camera pitch angle (from 0° to 90°) which impacts the object scale and viewpoint variation.
As illustrated in \cref{fig:3}\textcolor{iccvblue}{c}, the dataset spans an angle from 0° to 75°, achieving nearly full angular coverage excluding extreme situations.
This wide scope enriches the dataset with comprehensive multi-view object information. 

\textbf{Time statistics.}
It records the timestamp of image acquisition including year, month, day, hour, and minutes.
As shown in \cref{fig:3}\textcolor{iccvblue}{d}, the dataset spans a broad temporal range from dawn to night throughout all seasons, capturing various object characteristics in all-day and all-year conditions.

\textbf{Weather statistics.}
Textures in RGB images and thermal radiations in IR images are usually altered by varying weather conditions.
As seen in \cref{fig:3}\textcolor{iccvblue}{e}, the dataset contains 7 typical and extreme weather types, fostering improved detection availability in real-world applications.

\textbf{Illumination statistics.}
As noted in \cref{fig:3}\textcolor{iccvblue}{f}, images span 6 illumination levels from lightless to high-light.
Since object characteristics and image quality are sensitive to illumination especially in RGB modality, this diverse illumination boosts model\textquotesingle s robustness in real-world situations.

\textbf{Scenarios statistics.} As shown in \cref{fig:3}\textcolor{iccvblue}{g}, images were captured across 11 scenario types within Urban, Suburban, and Village, encompassing a wide range of environments such as Road, Neighborhood, Construction Site, Parking Lot, and so on.
High diversity of scenarios brings in complex interference from cluttered backgrounds.

\vspace{-0.1cm}
\subsection{Advances of ATR-UMOD Dataset}

\begin{figure*}[htb!]
  \centering
   \includegraphics[width=1.0\linewidth]{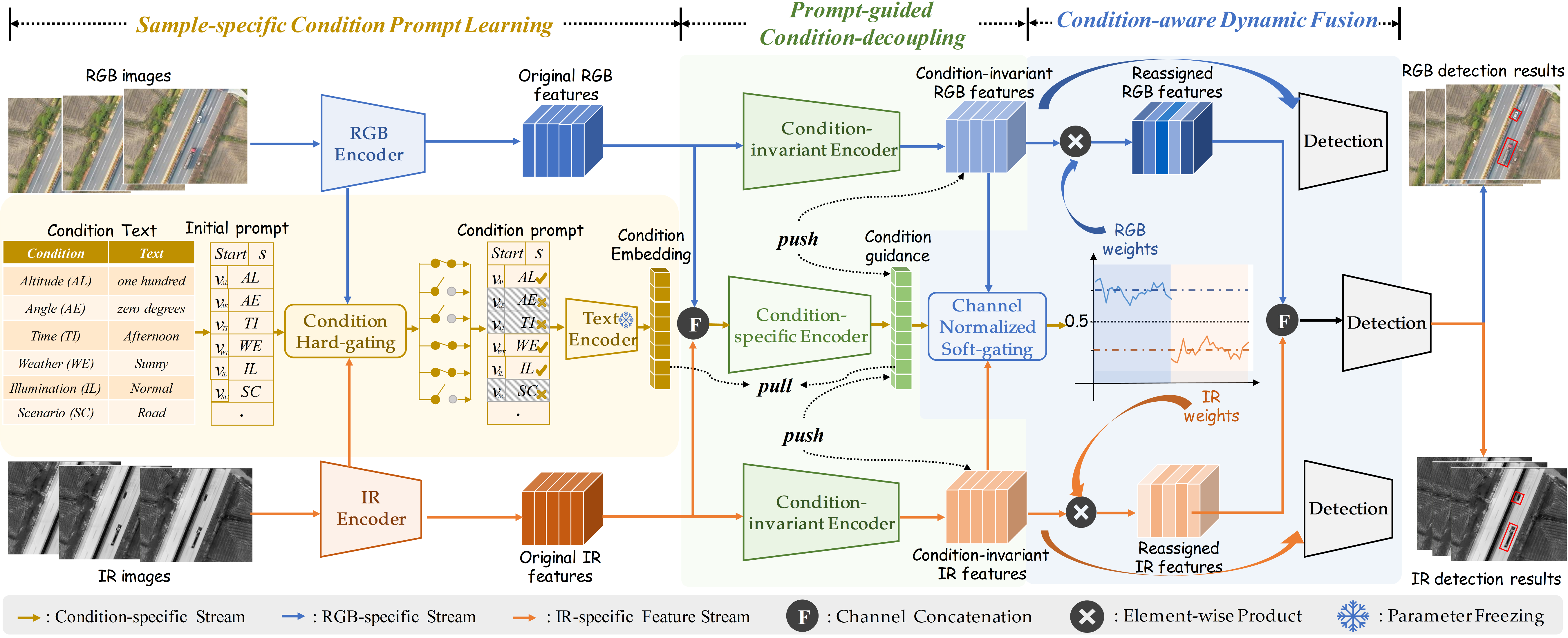}
   \vspace{-0.7cm}
   \caption{Overview structure of the proposed PCDF method. }
   \label{fig:4}
   \vspace{-0.7cm}
\end{figure*}

Compared with the existing RGB-IR UOD dataset, our ATR-UMOD has several unique advancements:

(1) \textbf{More diversified data distribution.}
Considering limited imaging conditions, our dataset significantly enhances condition diversity in several dimensions, including broader altitude ranges, extended angle coverages, comprehensive time span, challenging weather conditions, richer illumination variations, and more complex backgrounds.
These improvements allow the dataset to better reflect the complexity of real-world data distribution, making it a more comprehensive dataset for data-driven RGB-IR UOD.

(2) \textbf{Richer object types.}
The ATR-UMOD dataset contains 11 object categories, whereas the existing dataset is limited to 5 categories. The increased diversity of object type not only facilitates models in capturing subtle features but also enhances their ability to recognize a wider variety of targets for more complex real-world applications.


(3) \textbf{Extra condition information.}
Due to variations in multimodal image quality and object characteristics under different conditions, condition information is vital for the effectiveness of detection models.
To this end, ATR-UMOD first annotates 6 key condition attributes for each image pair, enabling deeper exploration of conditional impacts on multimodal object detection and making it a comprehensive benchmark for condition-specific performance evaluation.


\vspace{-0.2cm}
\section{Method}

\subsection{Overview}
\label{section:3.1}


Our method dynamically reassigns multimodal contributions based on multi-condition prompts.
As shown in \cref{fig:4}, RGB-IR image pairs are processed through dual-branch encoders to extract unimodal original features.
Simultaneously, condition texts are fed to SCPL to learn relevant condition prompts.
To address inaccessible conditions in practice, unimodal original features are decoupled into condition-specific and -invariant features.
The condition-specific features are aligned with condition embedding to gain condition guidance.
Multimodal weights are finally obtained by this guidance for dynamically reassigning contributions of the condition-irrelevant features for detection.
\subsection{Sample-specific Condition Prompt Learning}
\label{section:3.2}
Leveraging CLIP\textquotesingle s powerful text representation ability and rich textual information, we encode condition semantics through prompt learning.
However, different condition attributes affect individual samples to varying degrees~\cite{DBLP:conf/bmei/RenYGM19}, some may be negligible or even disruptive.
For example, scenario is often irrelevant under night illumination. 
Thus, using all attributes indiscriminately as reliability cues is unreasonable. 
To address this, SCPL learns relevant and effective attributes for each sample. 

\textbf{Initial prompt construction (IPC).} Given a set of condition attributes $\mathbb{A}  = \{\mathcal{A}_{1}, \mathcal{A}_{2}, \dots, \mathcal{A}_{N}\}$, where $N$ is the number of condition attributes and each attribute $\mathcal{A}_{n}$ comprises $M_{n}$ distinct classes, represented as $\mathcal{A}_{n} = \{a_{n}^{1}, a_{n}^{2}, \dots, a_{n}^{M_{n}}\}$, we create a initial condition prompt by formatting these attributes into a fixed template. This template comprises a subject description $s$ and several condition prefixs $v_{n}$. Details are provided in Supplementary material Sec. \textcolor{iccvblue}{B.1}.
Taken a condition prefix with a condition attribute as a condition block $\mathcal{O}_{i} = \{o_{1}^{i}, o_{2}^{i}, \dots, o_{N}^{i}\}$, the initial prompt $I_{i}$ for sample $i$ can be expressed as: 
\begin{equation}
  I_{i} = s+ \sum_{n=1}^{N} o_{n}^{i}, \quad o_{n}^{i} = v_{n} + a_{n}^{i}.
  \label{eq:1}
\end{equation}
The initial condition embedding $\mathcal{I}_{i}$ is obtained by feeding $I_{i}$ into the frozen text encoder of CLIP~\cite{DBLP:conf/icml/RadfordKHRGASAM21}.

\textbf{Sample-specific condition prompt-tuning (SCPT).}
To eliminate the effects of irrelevant attributes, we introduce a sample-specific prompt tuning mechanism based on hard-gating masks.
Inspired by that experts assess the influence of condition attributes on multimodal reliability by observing specific patterns in each sample~\cite{DBLP:journals/tcsv/LiuLZQ22}, we feed $\mathcal{I}_{i}$ with multimodal features into a condition hard-gating network to generate learnable sample-specific condition prompts.
Precisely, multimodal features $\mathcal{F}_{rgb}^{i}$, $\mathcal{F}_{ir}^{i} \in \mathbb{R}^{C \times H \times W}$ are fused through nonlinear layers $\boldsymbol{F}_{c}$ and a Softmax function $\sigma$ to yield attribute availability probabilities.
The hard-gating masks $\mathcal{G}_{i} = \{g_{1}^{i}, g_{2}^{i}, \dots, g_{N}^{i}\}$ are then obtained by indicator function $\mathbbm{1}$ with a predefined threshold $\tau$:
\begin{equation}
  \mathcal{G}_{i} = \mathbbm{1}(\sigma (\boldsymbol{F}_{c}(Pool(\mathcal{F}_{rgb}^{i}, \mathcal{F}_{ir}^{i}), \mathcal{I}_{i}))>= \tau),
\end{equation}
where $(\cdot,\cdot)$ is concatenation, Pool is maxpooling, and $\tau$ is set to 0.15 (see in Supplementary Sec. \textcolor{iccvblue}{G}).
\(g_{n}^{i} \in \{0, 1\}\) determines whether the \(n\)-th attribute should be included or excluded. The adjusted condition block \(o_{n}^{i'}\) is defined as:
\begin{equation}
o_{n}^{i'}\ = 
\begin{cases} 
o_{n}^{i}\, & \text{if } g_{n}^{i} = 1, \\
\emptyset, & \text{if } g_{n}^{i} = 0.
\end{cases}
\end{equation}

This gating mask is applied to the initial prompt to obtain the sample-specific condition prompt $P_{i}$:
\begin{equation}
  \begin{aligned}
    \vspace{-0.2cm}
    P_{i} = I_{i} \times \mathcal{G}_{i}= s+ \sum_{n=1}^{N} o_{n}^{i'}.
  \end{aligned}
  \vspace{-0.2cm}
\end{equation}

Finally, we transform $P_{i}$ into condition embeddings $\mathcal{F}_{t}^{i}$ with CLIP.
Noted that SCPL is only utilized in training.

\subsection{Prompt-guided Condition-decoupling}
\label{section:3.3}
As $\mathcal{F}_{t}^{i}$ is inaccessible in practice, condition guidance must be derived from visual features. 
However, directly modeling it from original features may bring in interferences between condition and object information.
Thus, we decouple original features into condition-specific and condition-invariant components, where the condition-specific features tied to condition semantics, while the condition-invariant features focus on robust object-discriminative representations.

To achieve this, we introduce a three-branch decoupling network.
Specifically, the first branch is the condition-specific encoder $\boldsymbol{S}$ that extracts condition-specific features $\mathcal{F}^{s,i}$ from the visual features.
Other branches consist of condition-invariant encoders $\boldsymbol{V}_{m}$ that independently extract condition-invariant features $\mathcal{F}_{m}^{v,i}$ from the unimodal features $\mathcal{F}_{m}^{i} \ (m \in \{rgb, ir\})$. This can be formulated as:
\begin{equation}
  \begin{aligned}
    \mathcal{F}^{s,i} = \boldsymbol{S}(F(\mathcal{F}_{rgb}^{i}, \mathcal{F}_{ir}^{i});\theta^{s}), \ \mathcal{F}_{m}^{v,i} = \boldsymbol{V}_{m}(\mathcal{F}_{m}^{i};\theta^{v}_{m}),
  \end{aligned}
\end{equation}
where $\theta_{m}^{v}$ and $\theta^{s}$ are the learnable parameters, $F(\cdot, \cdot)$ denotes the multimodal fusion function.

For $\mathcal{F}^{s,i}$, it is essential to ensure semantic consistency with the intended condition prompts $\mathcal{F}_{t}^{i}$. For this purpose, we adopt a prompt-guided distillation loss $L_{dt}$ to minimize the distance between the $\mathcal{F}^{s,i}$ and $\mathcal{F}_{t}^{i}$, which is defined by a widely used distance metric named CMD~\cite{DBLP:conf/iclr/ZellingerGLNS17}:
\vspace{-0.2cm}
\begin{equation}
  \begin{aligned}
  \mathcal{L}_{dt} =& \frac{1 }{\left | b-a \right |} \left \| \boldsymbol{E}(\mathcal{F}^{s,i})-\boldsymbol{E}(\mathcal{F}_{t}^{i}) \right \| _{2} \\
  +&\sum_{k=2}^{5} \frac{1 }{\left | b-a \right |^{k}} \left \| \boldsymbol{C}_{k}(\mathcal{F}^{s,i})-\boldsymbol{C}_{k}(\mathcal{F}_{t}^{i}) \right \|_{2} ,
  \vspace{-0.2cm}
\end{aligned}
\end{equation}
where $\boldsymbol{E}(\cdot)$ is the empirical expectation vector, $\boldsymbol{C}_{k}(\cdot)$ is the vector of $k$-th order sample central moments, and $\left [ a,b \right ]$ is the bound of the random variable $\mathcal{F}^{s,i}$ and $\mathcal{F}_{t}^{i}$.

For $\mathcal{F}_{m}^{v,i}$, the following properties must be satisfied: (1) it remains invariant to varying conditions; (2) it exhibits sufficient discrimination for effective object detection.
As condition guidance has been modeled from $\mathcal{F}^{s,i}$, we present a irrelevant loss $\mathcal{L}_{irr}$ for property (1) that highlights the dissimilarity between $\mathcal{F}_{m}^{v,i}$ and $\mathcal{F}^{s,i}$. It is achieved by the squared Frobenius norm $\left \| \cdot \right \| _{F}^2$:
\begin{equation}
  \begin{aligned}
    \mathcal{L}_{irr} =  \left\| \left( F_{rgb}^{v,i} \right)^{T} \mathcal{F}^{s,i} \right\|_F^2 + \left\| \left( F_{ir}^{v,i} \right)^{T} \mathcal{F}^{s,i} \right\|_F^2.
\end{aligned}
\end{equation}

For property (2), we introduce a discrimination loss $\mathcal{L}_{dc}$ by a detector to ensure the discriminative capacity of $\mathcal{F}_{m}^{v,i}$:
\begin{equation}
  \mathcal{L}_{dc} = \sum_{m\in{\{rgb, ir\}}} (\mathcal{L}_{cls}(F_{m}^{v,i}) + \mathcal{L}_{reg}(F_{m}^{v,i}) + \mathcal{L}_{obj}(F_{m}^{v,i})),
\end{equation}
where $\mathcal{L}_{cls}$, $\mathcal{L}_{reg}$, and $\mathcal{L}_{obj}$ are the classification, regression, and objectness loss, respectively.
Finally, the decoupling loss $\mathcal{L}_{dec}$ can be formulated as:
\begin{equation}
  \mathcal{L}_{dec} =\lambda_{1} \mathcal{L}_{dt} + \lambda_{2} \mathcal{L}_{irr} +\lambda_{3} \mathcal{L}_{dc},
\end{equation}
where $\lambda_i$ is the trade-off parameter that is experimentally set to 0.01, 0.003, and 0.01 respectively in this study.

\subsection{Condition-aware Dynamic Fusion}
\label{section:3.4}
The multimodal reliability is determined by the condition guidance $\mathcal{F}^{s,i}$.
Given that different channels capture distinct semantic aspects~\cite{DBLP:journals/ijcv/SelvarajuCDVPB20}, we introduce a channel-wise normalized soft-gating transformation to enhance model adaptability.
In detail, it adaptively maps $\mathcal{F}^{s,i}$ to multimodal weights $\mathcal{W}_{m}^{i} \in \mathbb{R}^{1 \times C}$ via a nonlinear projection function $\boldsymbol{F}_{t}$ followed by a channel-wise normalized operation, ensuring information preservation in fusion features while constraining weights within $[0, 1]$:

\vspace{-0.4cm}
\begin{equation}
  \begin{aligned}
    \mathcal{W}_{m}^{i} = \frac{\exp([\boldsymbol{F}_{t}(\mathcal{F}^{s,i})]_{m})}{ \exp([\boldsymbol{F}_{t}( \mathcal{F}^{s,i})]_{rgb})+\exp( [\boldsymbol{F}_{t}(\mathcal{F}^{s,i})]_{ir})},
\end{aligned}
\end{equation}
where $[\cdot]_{m}$ represents the channels of $m$ modality.
These weights are applied to \textit{condition-invariant features} $\mathcal{F}_{m}^{v,i}$ to dynamically adjust multimodal contributions.
Notably, only $\mathcal{F}_{m}^{v,i}$ are reassigned for the fusion process, mitigating interference of condition-induced noise.
The final fused feature $\mathcal{F}_{f}^{i}$ is obtained through a simple concatenation operation:
\begin{equation}
  \begin{aligned}
  \mathcal{F}_{f}^{i} = Concat(\mathcal{W}_{rgb}^{i} \odot \mathcal{F}_{rgb}^{v,i}, \mathcal{W}_{ir}^{i} \odot \mathcal{F}_{ir}^{v,i}),
\end{aligned}
\end{equation}
where $\odot$ denotes the element-wise multiplication. 
$\mathcal{F}_{f}^{i}$ is fed into detection head for task-oriented reliability learning.
This dynamic fusion adaptively leverages discriminative information from the dominant modality while suppressing contributions from the suboptimal one.

\vspace{-0.2cm}
\section{Experiments}
\label{sec:Experiments} 
\subsection{Implementation Details}
\label{section:4.1}

Our method was implemented in PyTorch on an NVIDIA RTX 4090 GPU. The network parameters were updated using SGD~\cite{DBLP:series/lncs/Bottou12} optimizer with an initial learning rate of 0.01 and decayed exponentially. Momentum and weight decay were set to 0.937 and 0.0005, respectively.
We utilized the ViT-B/16~\cite{DBLP:conf/iclr/DosovitskiyB0WZ21} pre-trained model from CLIP as a text encoder.
Our model comprises two trainable processes, including a fusion network with SCPL and the full pipeline, both trained for 50 epochs with an $640\times 512$ image size and a batch size of 16.
All baseline methods were trained with their original parameter settings to ensure optimal performance.
The Mean Average Precision (mAP) is adopted to evaluate the detection performance with an IoU of 0.5.

\begin{table*}[!t]\normalsize
  
    \renewcommand{\arraystretch}{1.05}
    \begin{center}
    \centering
    \tiny
    \setlength{\tabcolsep}{2.7mm}{
    \resizebox{1.0\linewidth}{!}{
    \begin{tabular}{ccccccccccccccc} 
      \bottomrule
      \rule{0pt}{6pt} 

    \textbf{Detectors}     & \makebox[0.001\textwidth][c]{\textbf{Modality} }     & \textbf{CR} & \textbf{SV} & \textbf{VN} & \textbf{BS} & \textbf{FC} & \textbf{TK}  & \textbf{TT} & \textbf{TR}  & \textbf{CE} & \textbf{ER}  & \textbf{ME} & \makebox[0.01\textwidth][c]{\textbf{mAP (\%) \textcolor[rgb]{1,0,0}{$\uparrow$}}}  \\ \hline 
    RetinaNet~\cite{DBLP:conf/iccv/LinGGHD17}     & \multirow{7}{*}{RGB}   &26.3 &29.9 &32.9 &59.6 &17.9 &23.1 &2.1          &4.7            &9.3                  &17.8       & 14.1        &21.6                       \\
    $\textup{S}^2$A-Net~\cite{DBLP:journals/tgrs/HanDLX22}   &  & 34.2 &44.9 &45.9 &69.2 &24.4             & 37.4                        &5.6        &  22.5         & 49.5              &31.2        &25.6         & 35.5                      \\
    Faster R-CNN~\cite{DBLP:journals/pami/RenHG017} & & 35.2 &49.0 &48.7 &69.4 &26.8 &44.0 &17.6 &35.0          &55.3           & 36.6                 &22.6                 & 40.0                       \\
    ReDet~\cite{DBLP:conf/cvpr/HanD0X21}    &  &36.9 &52.5 &51.6 &74.8 &33.5   & 48.1                        & 16.7         & 40.7          & 61.4                &  36.8       & 32.9      &  41.1                \\
    RoITransformer~\cite{DBLP:conf/cvpr/DingXLXL19}   &   &37.2 &53.3 &51.9 &71.5 &30.1   &46.8                        &18.2       &36.3       &58.9    & 38.3               &25.3          &42.5                   \\
    Oriented R-CNN~\cite{DBLP:conf/iccv/Xie0WYH21}  &  & 36.9 &52.5 &51.6 &74.8 &33.5      &48.1                         &16.7          &40.7            &  61.7              &36.8         & 32.9      &  44.2                  \\
    YOLOv5s~\cite{glenn_jocher_2020_4154370}   &       &45.8 &60.7 &57.5 &75.2 &41.6          &\textbf{52.1}                         &18.2          & 42.3           & \underline{68.7}               &  \underline{47.5}       &  47.4        & 50.7                       \\
    \hline
    RetinaNet~\cite{DBLP:conf/iccv/LinGGHD17}   & \multirow{7}{*}{IR}  &39.3 &29.1 &20.8 &48.9 &24.7 &13.1    &6.4          &1.1            &6.2                  &1.0       & 18.4      &18.9                 \\
    $\textup{S}^2$A-Net~\cite{DBLP:journals/tgrs/HanDLX22}     &            &50.2 &35.9 &31.8 &59.9 &35.5 &24.3    &31.4          &16.0            & 10.8                 & 1.0      &  32.0     &   29.9                \\
    Faster R-CNN~\cite{DBLP:journals/pami/RenHG017} &  &53.4 &39.0 &35.6 &64.9 &37.3 &28.0    &33.4          &25.7            &34.1                  &11.9       &  17.6     &  34.7            \\
    ReDet~\cite{DBLP:conf/cvpr/HanD0X21}   &   &57.4 &42.6 &38.8 &70.4 &42.3 &31.5    &52.0          & 15.9           &33.7                  &8.7       &23.1       &   37.9             \\
    RoITransformer~\cite{DBLP:conf/cvpr/DingXLXL19}  &   &54.6 &41.8 &38.7 &64.0 &43.1 &33.6    &61.0          &23.4            & 32.8                 & 7.0      &  23.4     &  38.5        \\
    Oriented R-CNN~\cite{DBLP:conf/iccv/Xie0WYH21}  &    &57.5 &41.6 &36.8 &63.8 &43.5 &28.6    & 64.3         &28.5            &44.2                  & 6.9      &   23.9    &  40.0             \\
    YOLOv5s~\cite{glenn_jocher_2020_4154370}  &     &65.8 &51.2 &51.6 &75.3 &53.1          &38.9                         & \underline{83.3}  &  \textbf{46.2}      &  57.9               & 12.0      &42.7         &52.5                   \\ 
    \hline
    IAF R-CNN~\cite{DBLP:journals/pr/LiSTT19} &\multirow{9}{*}{RGB+IR}  &51.9 &45.9 &48.3 &64.6 &37.8  &42.6  & 30.4         &30.8           &43.8               & 43.5      &20.8    &41.9              \\
    Halfway Fusion~\cite{DBLP:conf/bmvc/LiuZWM16} &  &53.1 &47.0 &51.3 &73.5 &42.1  &42.4  &39.5          &34.4           &52.5               &  35.3      & 22.9   & 44.9             \\
    UA-CMDet~\cite{DBLP:journals/tcsv/SunCZH22}  &    &50.9 &43.3 &47.9 &75.8 &51.4  &44.5 &42.8 &40.1          &54.8           &39.6               &23.2        &46.8                        \\


    $\textup{C}^2$Former~\cite{DBLP:journals/tgrs/YuanW24} &   &60.5 &53.3 &51.6 &81.6 &46.1        & 44.7                        &46.6   & 29.3           &   56.3         & 36.8         & 40.0        &  49.7                \\
    TINet~\cite{DBLP:journals/tim/ZhangYHWY23} &&60.2  &51.4 &54.4 &74.5 &50.2 &46.0  &44.6  &39.7          &59.0           &\underline{47.5}               &27.0       &50.4                  \\
    CALNet~\cite{DBLP:conf/mm/HeTZZ23}      &   &\underline{71.9} &\textbf{65.5} &\textbf{71.0} &78.4 &53.6        &51.2                         &37.7   &  35.3          &56.3            &31.9          &38.6         &53.8                    \\ 

    OAFA~\cite{DBLP:conf/cvpr/ChenQLBFHZ24} &   &70.4 &59.6 &63.1 &81.5 &60.1        & 47.5                        &80.1   &  32.4          &59.0            &33.0          &50.1         &57.9             \\
    YOLOrs~\cite{DBLP:journals/staeors/SharmaDKCPMS21} &   &\textbf{73.2} &\underline{62.6} &\underline{66.3} &\underline{81.8} &\underline{61.1}        & 48.2                        &70.3   &  37.6          &64.3            &41.8          &\textbf{52.9}         &\underline{60.0}          \\
    
    \rowcolor{gray!20}
    PCDF (Ours) &    &70.8 &60.6 &65.4 &\textbf{84.3} &\textbf{62.1} &\underline{51.3} &\textbf{86.1}        &\underline{42.5}       & \textbf{71.1}             &\textbf{49.0}         &\underline{51.2}      &\textbf{63.1}          \\ \bottomrule
    \end{tabular}
    }}

    \vspace{-0.35cm}
    \caption{Detection results (in \%) on the ATR-UMOD dataset. All detectors perform object localization and classification with OBB heads. Best results are marked with \textbf{bold}, while the second one is highlighted in \underline{underline}.}
    \label{tab:2}
    \end{center}
    \setlength{\abovecaptionskip}{0.cm} 
    \vspace{-1.0cm}
    \end{table*}

\subsection{Results Comparisons}
\label{section:4.2}
We evaluate PCDF on the ATR-UMOD dataset through comprehensive qualitative and quantitative analyses, benchmarking against 7 state-of-the-art (SOTA) \textit{unimodal detectors}, including RetinaNet~\cite{DBLP:conf/iccv/LinGGHD17}, $\textup{S}^2$A-Net~\cite{DBLP:journals/tgrs/HanDLX22}, Faster R-CNN~\cite{DBLP:journals/pami/RenHG017}, ReDet~\cite{DBLP:conf/cvpr/HanD0X21}, RoITransformer~\cite{DBLP:conf/cvpr/DingXLXL19}, Oriented R-CNN~\cite{DBLP:conf/iccv/Xie0WYH21}, and YOLOv5s~\cite{glenn_jocher_2020_4154370}, as well as 8 \textit{multimodal detectors}, including IAF R-CNN~\cite{DBLP:journals/pr/LiSTT19}, Halfway Fusion~\cite{DBLP:conf/bmvc/LiuZWM16}, UA-CMDet~\cite{DBLP:journals/tcsv/SunCZH22}, $\textup{C}^2$Former~\cite{DBLP:journals/tgrs/YuanW24}, TINet~\cite{DBLP:journals/tim/ZhangYHWY23}, CALNet~\cite{DBLP:conf/mm/HeTZZ23}, OAFA~\cite{DBLP:conf/cvpr/ChenQLBFHZ24}, and YOLOrs~\cite{DBLP:journals/staeors/SharmaDKCPMS21}.
Among these methods, IAF R-CNN and TINet are illumination-guided fusion methods.
Our baseline is a one-stage dual-stream detector that integrates two modalities through concatenation fusion.
Noted that the multimodal detectors are all trained with IR labels.

\textbf{Quantitative comparison.} The quantitative comparisons are presented in \cref{tab:2}.
The mAP results demonstrate that PCDF significantly outperforms the SOTA unimodal and multimodal methods, surpassing the second-best method by 3.1\%.
Moreover, PCDF consistently excels across multiple categories, achieving the best or second-best performance in most cases while maintaining competitive results in the remaining ones.
This demonstrates the effectiveness of our approach in dynamically leveraging reliable information from both RGB and IR modalities, thereby enhancing detection performance.

\begin{table}[!t]\normalsize
  \tiny
  \renewcommand{\arraystretch}{1.0}
  \begin{center}
  \centering
  
  \setlength{\tabcolsep}{2.2mm}{
  \resizebox{1.0\linewidth}{!}{
  \begin{tabular}{|cc|c|c|c|c|} 
    \hline
    
    \multicolumn{2}{|c|}{\multirow{2}{*}{\textbf{Conditions}}}       &\multicolumn{4}{c|}{\textbf{Method}} \\\cline{3-6}
        &           & \textbf{CALNet}      & \textbf{OAFA} & \textbf{YOLOrs}   & \textbf{Ours}  \\ \hline 
    
      \multicolumn{1}{|c}{\multirow{2}{*}{AL}}   & \multicolumn{1}{|c|}{$\left[0,120\right]$}          &55.3     &  59.4    &  61.8    & \textbf{67.0}                               \\\cline{2-6}
      &\multicolumn{1}{|c|}{$\left( 120, 300 \right]$}  & 41.4 &47.3  & 48.3     & \textbf{50.3}                               \\  \hline\hline

      \multicolumn{1}{|c}{\multirow{2}{*}{AN}}   & \multicolumn{1}{|c|}{$\left[0,30\right]$ }       &51.0    & 58.2 &58.5      & \textbf{60.4 }                              \\\cline{2-6}
            &\multicolumn{1}{|c|}{$\left(30,75\right]$}  &43.1 &51.7  &55.0 & \textbf{57.2}                             
                \\ \hline\hline

      \multicolumn{1}{|c}{\multirow{3}{*}{TI}} &\multicolumn{1}{|c|}{Morning}  &43.9   & 53.6  &55.6   & \textbf{58.9}                              \\\cline{2-6}
         & \multicolumn{1}{|c|}{Afternoon }           &53.8     & 60.0     &62.7      & \textbf{66.3}                               \\\cline{2-6}
      &\multicolumn{1}{|c|}{Night }       &31.2          & 38.2  & 35.3      & \textbf{41.7}                              \\  \hline\hline
      
      \multicolumn{1}{|c}{\multirow{5}{*}{WE}}   & \multicolumn{1}{|c|}{Cloudy} &54.8  & 62.1     &69.6      & \textbf{71.2}                \\\cline{2-6}
      &\multicolumn{1}{|c|}{Foggy}  &28.8   &30.9  &\textbf{33.2}   & 32.6                               \\\cline{2-6}
            & \multicolumn{1}{|c|}{Snowy}        &38.2    & 48.4  & 50.1     &  \textbf{53.4}                       \\\cline{2-6}
            & \multicolumn{1}{|c|}{Sunny  }     &48.7          & 54.6  & 57.1     &  \textbf{60.1}                       \\\cline{2-6}
            & \multicolumn{1}{|c|}{Night  }     &32.5      & 38.0  & 35.4     &  \textbf{41.3}                       \\ \hline\hline
            \multicolumn{1}{|c}{\multirow{4}{*}{IL}}   & \multicolumn{1}{|c|}{Normal}    &49.8   & 55.0     & 56.9     & \textbf{60.6}                               \\\cline{2-6}
            & \multicolumn{1}{|c|}{Dim}    &52.5  & 59.8     & 67.5     & \textbf{67.7}                               \\\cline{2-6}
            & \multicolumn{1}{|c|}{Near Night}   & 41.5  & 44.8    &   47.5   & \textbf{48.1}                               \\\cline{2-6}
      &\multicolumn{1}{|c|}{Night}   &31.5  &37.3   &34.6    & \textbf{40.0 }                               \\ 
      \hline\hline
      \multicolumn{1}{|c}{\multirow{5}{*}{SC}}   & \multicolumn{1}{|c|}{Construction site}  &41.0 &  42.3    & 47.2     & \textbf{49.0}                               \\\cline{2-6}
      &\multicolumn{1}{|c|}{Factory}  &17.6  & \textbf{23.0}  & 22.4& \textbf{23.0}                               \\\cline{2-6}
      & \multicolumn{1}{|c|}{Neighborhood }      &33.7         & 34.6  & 31.9      & \textbf{35.3}                              \\\cline{2-6}
      & \multicolumn{1}{|c|}{Parking Lot }     &31.9        & 39.5  &38.0       & \textbf{40.8}                              \\\cline{2-6}
      & \multicolumn{1}{|c|}{Road }        &49.6       & 57.0  &  59.3     & \textbf{62.3}                              \\ \hline
  \end{tabular}
  }}
\end{center}
\vspace{-0.6cm}
\caption{Detection results (in \%) in different conditions. Noted that AL, AN, TI, WE, IL, and SC represent altitude, angle, time, weather, illumination, and scenario, respectively. 
  Best results are marked with \textbf{bold}.}
  \label{table:6}

\vspace{-0.9cm}
\end{table}

\begin{figure*}[htb!]
  \centering
   \includegraphics[width=0.99\linewidth]{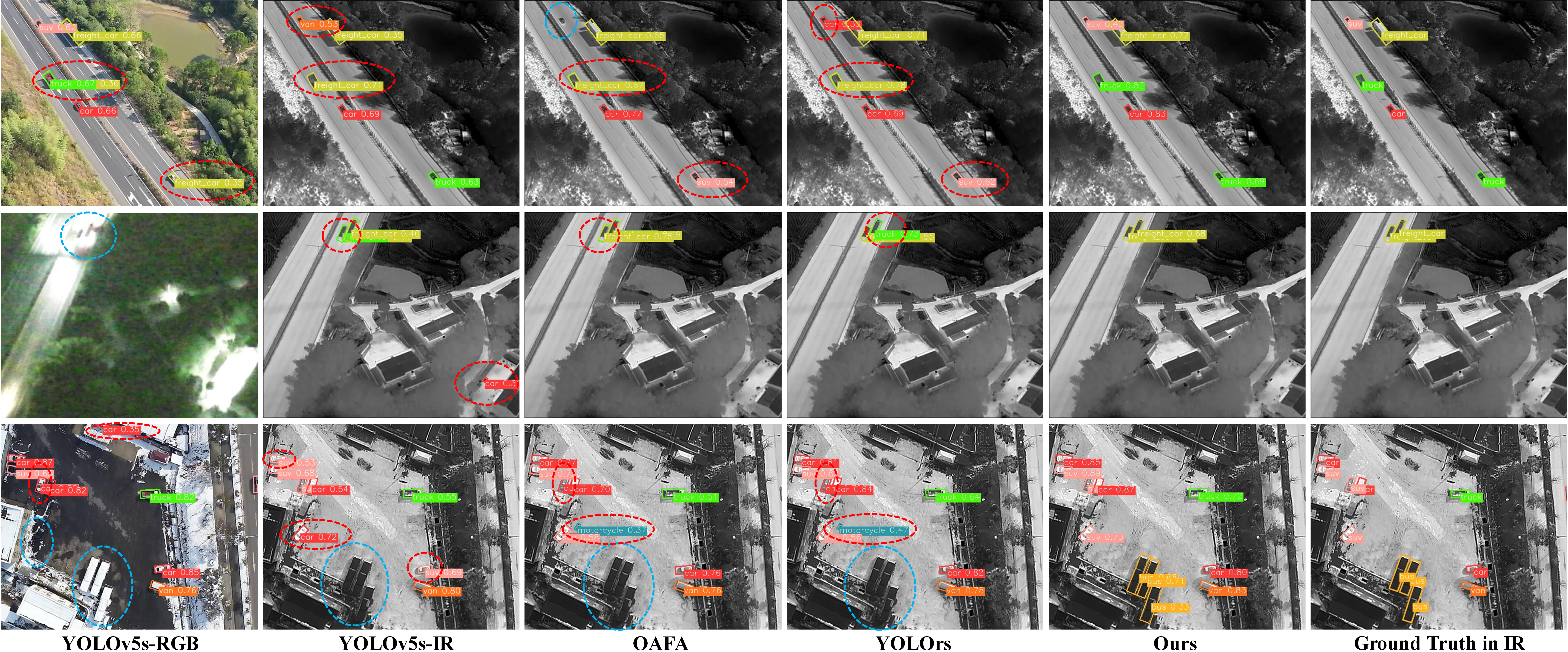}
   \vspace{-0.2cm}
   \caption{Qualitative comparison on ATR-UMOD dataset. Confidence threshold is 0.25. Fusion method results are displayed on IR images to align with the supervisory labels. \textcolor{cyan}{Missed} and \textcolor{red}{incorrectly} detected objects are indicated with \textcolor{cyan}{blue} and \textcolor{red}{red} dashed circles, respectivelely.}
   \label{fig:5}
   \vspace{-0.7cm}
\end{figure*}

\textbf{Qualitative comparison.} \cref{fig:5} provides qualitative comparisons across typical conditions among the SOTA unimodal model, SOTA multimodal models, and PCDF.
In Overexposure (first row), Night (second row), and Snowy (third row) conditions, RGB and IR modalities exhibit distinct reliabilities.
Unimodal methods struggle with excessive exposure, low visibility, and occlusion in RGB images, as well as insufficient information in IR images, leading to detection failures.
Fusion methods also fail to handle these challenging conditions effectively due to their rigid fusion strategies.
In contrast, our method dynamically exploits the complementarity of RGB and IR modalities, achieving superior detection performance.


\subsection{Results on Different Conditions}

To assess the effectiveness of our method across varying conditions, we conduct comprehensive experiments on ATR-UMOD dataset.
\cref{table:6} presents the detection results under different conditions in the SOTA multimodal methods and our PCDF.
Due to the excessive number of conditions, sample size for each condition was often insufficient, causing overfitting and impairing model training.
Therefore, conditions were appropriately merged in \cref{table:6}.
Details are privided in Supplementary material Sec. \textcolor{iccvblue}{B.2}.
The results indicate that PCDF achieves superior performance across nearly all conditions, demonstrating its robustness and adaptability in diverse conditions.
The suboptimal performance in ``Foggy'' condition may be attributed to inconsistencies in fog levels and visibility, which can be better addressed through fine classification in future work.

\subsection{Ablation Study}
\label{section:4.5}  
\vspace{-0.2cm}

\textbf{Effectiveness of SCPL}.
This module is designed to adaptively construct effective prompts with relevant conditions. To assess its impact, we conduct two ablation experiments. 
(1) \textit{w/o SCPT}: prompts are constructed solely using the initial prompts.
The performance drop in \cref{tab:module_performance} suggests that without SCPT, model captures unnecessary condition semantics while diluting the influence of the meaningful ones, leading to unreliable condition representations.
(2) \textit{w/o SCPL}: since SCPL is the foundation of PCD, we replace condition guidance with data guidance by applying channel attention~\cite{DBLP:conf/cvpr/HuSS18} for dynamic fuison.
The performance improvement over the baseline highlights the significance of dynamic fusion.
However, its performance is still inferior to PCDF, underscoring the essential role of condition-based information in mitigating multimodal reliance bias.

\textbf{Effectiveness of PCD}. 
It enables PCDF to test with condition information without condition labels.
\cref{tab:module_performance} shows a mAP drop of 1.2\% when PCD is removed. The reason lies in that PCD mitigates condition-induced noise interference by decoupling condition-irrelevant features, improving generalization across varying conditions.
Moreover, w/o $L_{dt}$, $L_{irr}$, or $L_{dc}$ in PCD also result in varying degrees of performance degradation, underscoring their roles in maintaining semantic consistency between the condition guidance and condition-specific features, separating condition-irrelevant and specific features, and enhancing the discriminability of condition-irrelevant features, respectively.

\textbf{Effectiveness of CDF}.
It aims to dynamically reassign multimodal contributions in response to condition variations.
Ablation studies were conducted by replacing CDF with simple fusion that integrate condition features into multimodal visual features via addition or concatenation.
The results reveal a notable performance decline, which can be attributed to the lack of direct relationship perception between conditions and multimodal contributions while introducing condition noise into the fusion process.

\begin{table}[htbp]
  \vspace{-0.3cm}
  \centering
  \tiny
  \setlength{\tabcolsep}{3.7mm}{
  \resizebox{1.0\linewidth}{!}{
  \begin{tabular}{ccc}
  \toprule
  \textbf{Module Name} & \textbf{Experimental Design} & \textbf{{{mAP (\%) \textcolor[rgb]{1,0,0}{$\uparrow$}}}}   \\
  \midrule
  Baseline & N/A & 58.4 \\
  \midrule
  \multirow{2}{*}{SCPL} & w/o SCPT &62.3   \\
   & w/o SCPL & 60.5\\
  \midrule
  \multirow{4}{*}{PCD} & w/o $\mathcal{L}_{dt}$ & 61.6  \\
   & w/o $\mathcal{L}_{irr}$ &  62.7\\
   & w/o $\mathcal{L}_{dc}$ & 62.0 \\
   & w/o PCD & 62.1  \\
  \midrule
  \multirow{2}{*}{CDF} & w/o CDF (add) &62.0  \\
  & w/o CDF (concat) &61.5  \\
  \midrule
  PCDF & N/A &63.1  \\
  \bottomrule
  \end{tabular}}}
  \vspace{-0.3cm}
  \caption{Ablation study on PCDF. ``w/o'' means without.}
  \label{tab:module_performance}
  \vspace{-0.7cm}
  \end{table}

\vspace{-0.2cm}
\section{Conclusion}
\label{sec:Conclusion}
\vspace{-0.2cm}
In this paper, we built a high-diversity RGB-IR UOD dataset featuring fine-grained object types, broad altitude ranges, extensive angle coverage, comprehensive time span, challenging weather conditions, rich illumination variations, diverse scenario types, and additional condition annotations.
Recognizing visual information bottlenecks in such diverse conditions, we incorporate conditions as contextual prompts for dynamically reassigning multimodal features.
Leveraging CLIP\textquotesingle s powerful semantic representations, we construct sample-specific condition prompts and design a soft-gating transformation to establish task-specific relationships between prompts and multimodal contributions.
A condition-decoupling mechanism enables testing without condition annotations.
Experiments on ATR-UMOD dataset validate the SOTA performance of our method.

\noindent \textbf{Acknowledgements.} This work was supported by the National Natural Science Foundation of China (62201586).

{
    \small
    \bibliographystyle{ieeenat_fullname}
    \bibliography{main}
}

\end{document}